\documentclass{article}

\usepackage{microtype}
\usepackage{graphicx}
\usepackage{booktabs} 

\usepackage[accepted]{icml2023}

\usepackage{amsmath}
\usepackage{amssymb}
\usepackage{mathtools}
\usepackage{amsthm}
\usepackage{tikz}
\usepackage[edges]{forest}
\usepackage{tabularx}
\usepackage[hyphenbreaks]{breakurl}

\usepackage[capitalize,noabbrev]{cleveref}

\theoremstyle{plain}

\theoremstyle{definition}

\theoremstyle{remark}

\usepackage[textsize=tiny]{todonotes}

\usetikzlibrary{shadows}
\usetikzlibrary{cd}

\graphicspath{ {./figures/} }

\icmltitlerunning{Neurosymbolic AI on Biomedical KGs}

\begin{document}

\twocolumn[
\icmltitle{Neurosymbolic AI for Reasoning on Biomedical Knowledge Graphs}



%
\icmlsetsymbol{equal}{*}

\begin{icmlauthorlist}
\icmlauthor{Lauren Nicole DeLong}{edi},
\icmlauthor{Ramon Fern\'{a}ndez Mir}{edi},
\icmlauthor{Zonglin Ji}{edi},
\icmlauthor{Fiona Niamh Coulter Smith}{edi},
\icmlauthor{Jacques D. Fleuriot}{edi}
\end{icmlauthorlist}

\icmlaffiliation{edi}{Artificial Intelligence and its Applications Institute, School of Informatics, University of Edinburgh, Edinburgh, UK}

\icmlcorrespondingauthor{Lauren Nicole DeLong}{L.N.DELONG@sms.ed.ac.uk}

\icmlkeywords{Neurosymbolic AI, knowledge graphs, bimoedicine, representation learning, hybrid AI, graph neural networks.}

\vskip 0.3in
]



\printAffiliationsAndNotice{}  

\begin{abstract}

Biomedical datasets are often modeled as knowledge graphs (KGs) because they capture the multi-relational, heterogeneous, and dynamic natures of biomedical systems. KG completion (KGC), can, therefore, help researchers make predictions to inform tasks like drug repositioning. While previous approaches for KGC were either rule-based or embedding-based, hybrid approaches based on neurosymbolic artificial intelligence are becoming more popular. Many of these methods possess unique characteristics which make them even better suited toward biomedical challenges. Here, we survey such approaches with an emphasis on their utilities and prospective benefits for biomedicine.

\end{abstract}


\section{Introduction}

Often, biomedical concepts can be understood as part of a system comprising highly multi-relational, heterogeneous, and interconnected entities. For instance, molecular interactions are observed to engage in dynamic, condition-dependent networks \cite{chang2001mammalian}. Graph structures are a popular choice for representing biomedical systems as they facilitate the storage of such data \cite{li2022graph}.

Recently, several artificial intelligence (AI) and machine learning (ML) approaches for analyzing biomedical data \cite{esteva2017dermatologist, savoy2020idx} and assisting clinical treatment planning \cite{wang2019artificial} have been reported. Hence, we anticipate that AI will become an integral part of modern healthcare processes. As this becomes more of a reality, considering the ethical and practical benefits of model interpretability, the inherent ability to understand how and why predictions were made, could significantly improve the resultant quality of care. However, facilitating model interpretability is a challenge as state-of-the-art methods tend to lack it \cite{chakraborty2017interpretability}. Furthermore, while patterned domain knowledge between general concepts in biology is ample \cite{ashburner2000gene, guvench2016computational}, few data-driven ML approaches explicitly encode the high-level relationships from those knowledge bases. 

Recently, however, neurosymbolic methods have managed to leverage the benefits of both rule-based inference and deep learning. These methods possess unique characteristics which are suited to address the aforementioned challenges. To facilitate a better understanding of these characteristics, we survey several of such approaches that were used on biomedical graphs. In our previous survey \cite{delong2023neurosymbolic}, we outlined three major taxonomic categories for neurosymbolic reasoning on graph structures:

\begin{enumerate}
    \item logically-informed embedding approaches,
    \item embedding approaches with logical constraints, and
    \item rule-learning approaches.
\end{enumerate}

Using these categories as guides, we classify and compare the approaches with a novel concentration on their utilities toward biomedical challenges. Specifically, we focus on and refer back to the interpretability, incorporation of domain knowledge, and sparsity handling abilities of each approach since these traits are particularly interesting for biomedical applications (Section \ref{sub:nesyai}). Additionally, we propose several prospective directions within biology and medicine which could benefit from this area of work.

\section{Background}

\subsection{Knowledge Graphs}

\noindent A \emph{graph} or \emph{network} structure is composed of a finite set of vertices or nodes, \emph{V}, and a finite set of edges, \textit{E}, connecting the vertices. Each edge $e = (u, v) \in E$ connects two \textit{adjacent} or \emph{neighbouring} vertices, creating a \textit{triple} \cite{rahman2017basic}. A \emph{knowledge graph} (KG), in particular, uses a graph structure to represent a knowledge base for some specific domain. 

KGs are a practical way to represent datasets as they capture multi-relational data well, are fast and easy to query, and allow us to store many different types of data in a similar format (a ``universal language'') \cite{nickel2015review, ehrlinger2016towards}. Notably, KGs can represent growing or changing knowledge bases effectively as they permit flexible storage schemas \cite{vicknair2010comparison}. As biomedical systems are diverse, tightly interconnected, complex, and frequently updated, the KG structure is a popular choice to store biomedical knowledge \cite{li2022graph}.

\subsection{Knowledge Graph Completion}
\label{sub:dichotomy}

Many KGs are inherently incomplete as they are limited to the knowledge we, as humans, possess \cite{chen2020knowledge}. Biomedical KGs are no exception; our knowledge of biological macromolecules and their interactions is still only a fraction of what is predicted to exist \cite{povolotskaya2010sequence, jiang2014impact}. Therefore, \textit{KG completion (KGC)}, the practice of adding new information to a graph, is a popular category of reasoning tasks on KGs. One prominent form of KGC is \textit{link prediction}, where the goal is to predict whether an edge exists between two nodes \cite{zitnik2018modeling, chen2020knowledge, zhang2021drug}. Here, we contrast symbolic methods for KGC (particularly link prediction) against corresponding embedding-based methods.

\subsubsection{Symbolic Methods for KGC}
\label{subsub:symbolic}

Some of the simplest and most intuitive methods toward KGC utilize a set of rules, which often use logical inference to perform link prediction. For example, consider the following background knowledge regarding the three molecular entities, Raf, MEK, and ERK, used in a common cell signaling mechanism \cite{chang2003signal}:
\begin{equation*}
\small\text{activates}(Raf, MEK),
\end{equation*}
\begin{equation*}
\small\text{activates}(MEK, ERK),
\end{equation*}
\begin{equation*}
\small\text{increases\_activity}(X, Z) \leftarrow \text{activates}(X, Y) \land \text{activates}(Y, Z).
\end{equation*}
From this, we deduce that $\text{increases\_activity}(Raf, ERK)$. The source of said rules varies between approaches. Ontologies, for example, are formalisations of knowledge base semantics which often represent the unique patterns and hierarchies within a specific knowledge domain \cite{dou2015semantic}. From such patterns, one can generate rules to infer new knowledge. Ontologies, such as those discussed in Section \ref{sub:biology}, are ample in biomedicine \cite{ashburner2000gene, schriml2022human}. However, they do not exist for every domain, and existing ones sometimes impose limitations on the type of predictions made. Alternatively, other methods mine rules directly from the KG \cite{galarraga2013amie}. In these cases, rules are based on the association patterns that are found to exist within the KG rather than within the application domain.

The major benefit of rule-based approaches is that they are inherently interpretable. While interpretability is defined in various ways across the literature, it is generally viewed as the ability to be understood by a human \cite{marcinkevivcs2020interpretability}. For instance, one could refer back to the rules which governed the algorithm to get a human-readable understanding of how and why certain predictions were made. Unfortunately, compared to other methodologies, such as those discussed in the next section, rule-based methods do not always achieve the best performance \cite{gema2023knowledge}. Additionally, when faced with large KGs, they often suffer from scalability issues such as exponential computational complexity \cite{zhou2021progresses}. This is an increasingly pressing issue with the rise of big data and digitalization \cite{ji2021survey}.

\subsubsection{Embedding Methods for KGC}
\label{subsub:kge}

In contrast to rule-based methods, KG embedding (KGE) methods often scale well to large datasets. In general, such methods aim to compute embeddings, low-dimensional vector representations of the graph constituents, so that similarity in the embedding space should approximate some sort of similarity in the original graph \cite{zhang2018link, li2020network}. They have recently increased in popularity due to their competitive performances in fields such as biochemistry \cite{zitnik2018modeling} or disease state prediction \cite{ravindra2020disease}. While some KGE methods use a pre-determined function to generate embeddings, others, such as a Graph Neural Network (GNN) use neural networks and deep learning to learn the best function. 

Despite their competitive performances, KGE approaches are often black-box models, so they lack interpretability \cite{chakraborty2017interpretability}. This can be problematic ethically and practically in areas like medicine, as an end-user cannot see whether predictions are reliably generated or understand the grounds that justify novel predictions \cite{norori2021addressing, olejarczyk2021patient}. 

\section{Biomedical Concepts and KGs}

To appreciate the usefulness of methodological differences between the surveyed approaches, it is critical to understand their shared biomedical goals and the corresponding datasets on which they operated. In the following sections, we explore what these KGs comprise and explain the biomedical prediction tasks addressed. 

\subsection{Common Biomedical Concepts}
\label{sub:biology}

To understand the concepts modeled within biomedical KGs, we briefly describe the most common biomedical terms used within the literature. Fields of biology are organized on varying levels of complexity. From simplest (or smallest) to most complex, biology is studied on the molecular, cellular, tissue, anatomical, individual, or population levels, among others \cite{motofei2022biology, garland2022trade}. Generally, one can understand each subsequent level as a system which envelopes the previous level. While clinical studies that focus on populations of patients often translate well to treatment on the individual level, studies on the former levels, especially the molecular level, help researchers uncover disease etiologies and regular bodily processes through the mechanisms and interactions of the body's most elementary units \cite{funkhouser2020pathology}.

Here, we broadly define key biomedical concepts. \textit{Genes} are pieces of genetic material (DNA) which instruct the cell as to which functional products should be made. \textit{Proteins}, such as Raf, MEK, and ERK introduced in Section \ref{subsub:symbolic}, are the resultant functional products that perform important activities within the cells \cite{cooper2022cell}. Together, these activities work to create \textit{biological processes (BPs)} \cite{ashburner2000gene}. \textit{Compounds} typically describe small chemicals found naturally or supplemented. In contrast, \textit{drugs} are compounds introduced into the body deliberately to interact with BPs and achieve some modified, often medicinal, effect. Compounds and drugs are sometimes referred to interchangeably throughout the literature \cite{tatonetti2012offsides, coleman2016adverse}. A \textit{drug indication} describes the condition that a drug is intended to treat, while a \textit{side effect} is an unintended and often adverse effect induced by a drug \cite{coleman2016adverse}. These components interact in a complex and highly dynamic fashion. Different individuals vary in genetic makeup, and within individuals, various tissues and cell types carry out distinct BPs. To accomplish those unique BPs, therefore, different cell types express, or read, different genes to create specific proteins. Because the proteins generated vary across cell and tissue types, the ways in which compounds interact vary across those types, leading to altered indications and side effects \cite{mauger2019mrna, buccitelli2020mrnas, ross2021proteome}.

\subsection{Biomedical KGs Used}
\label{subsub:kgs}

All of the approaches surveyed here were used on a biomedical KG. Notably, many of them used a particular KG containing information from the unified medical language system (UMLS) \cite{mccray2003upper, bodenreider2004unified} because it is openly available and a commonly used benchmark dataset \cite{kok2007statistical, dai2020survey}. Other approaches deviate from this trend, however, and choose a KG more specific to biomedical research goals. All openly available biomedical KGs mentioned in this survey are summarized in Table \ref{tab:kgs}.

\begin{table*}[t!]
\centering
\small
\caption{Knowledge Graphs Commonly Used Across Surveyed Approaches}
\label{tab:kgs}
\begin{tabular}{
|p{0.1\linewidth}
|p{0.3\linewidth}
|p{0.06\linewidth}
|p{0.04\linewidth}
|p{0.08\linewidth}
|p{0.04\linewidth}|p{0.2\linewidth}|} \toprule 
\textbf{KG} & \textbf{Origin} & \textbf{Nodes} & \textbf{Node Types} & \textbf{Triples}   & \textbf{Edge Types} & \textbf{Comprises} \\ \midrule 

UMLS                &  \url{https://www.nlm.nih.gov/research/umls/index.html}         & 893,025        & 135                     & 5,960               & 49                      & ontology-derived biomedical concepts\\ \midrule 

SNAP &  \url{https://github.com/williamleif/graphqembed} & 97k            &  UNK & \textgreater 8 mil. & 42                      & drugs, diseases, proteins, side effects, and BPs             \\ \midrule 

kg-covid-19 &  \url{https://github.com/Knowledge-Graph-Hub/kg-covid-19}       & 574,778        & 20                      & 24,145,562          & 41                      & drug, protein, gene interactions relevant to COVID-19\\ \midrule 

Hetionet    &  \url{https://github.com/hetio/hetionet}           & 47,031         & 11                      & 2,250,197           & 24                      & genes, compounds, and diseases       \\ \midrule 

OREGANO     &  \url{https://gitub.u-bordeaux.fr/erias/oregano}   & 152,086        & 10                      & 680,757             & 14                      & drugs, proteins, genes, and diseases \\
\bottomrule 
\end{tabular}
\end{table*}

\subsection{Biomedical Prediction Tasks}

For the most part, studies which use the UMLS KG use it as a benchmark dataset. In contrast, studies with a more specific biomedical research goal tend to use other KGs better tailored toward that challenge. For example, because the COVID-19 pandemic accelerated biomedical publication output by $\sim$10 000 new articles per month \cite{chen2023litcovid}, the creators of \textbf{BioGRER} (Section \ref{subsub:iterative}) \cite{zhao2020biomedical} aim to refine a COVID-19 KG built by automatic information extraction from literature. While KG refinement is not unique to biomedicine, it is highly important, especially after the surge of pandemic-based literature.

A more prominent challenge in ML-based biomedical studies, including those within this survey, is \textit{drug repositioning, or repurposing (DR)}. DR is the use of a clinically approved drug to treat another condition for which it was not originally intended \cite{ashburn2004drug}. Because a drug must pass expensive and time-consuming clinical trials before it gains clinical approval, using an approved drug for a novel indication saves time and money \cite{jayasundara2019clintrialstats}. This process was especially important for finding potential therapies quickly during the COVID-19 pandemic \cite{schultz2023machine}. DR is often formatted as a link prediction task between a node representing a drug and another representing some condition or treatment \cite{schultz2023machine}.

As the DR challenge has existed for several decades \cite{ashburn2004drug}, approaches varying from ontology-derived semantic networks to deep learning methods have been developed to address it \cite{xue2018review, luo2021biomedical}. Since approaches to DR already reflect the previously described methodological dichotomy from Section \ref{sub:dichotomy}, it is an ideal area to develop neurosymbolic, hybrid approaches. Furthermore, as Luo \textit{et al.} point out, many previous approaches do not address the sparsity of verified indication edges, so the integration of biomedical domain knowledge into state-of-the-art models could ameliorate that issue \cite{luo2021biomedical}. Finally, since molecular interactions are both spatially and temporally dynamic \cite{mauger2019mrna, buccitelli2020mrnas, ross2021proteome}, drug indications discovered in one cell or tissue model do not always translate to others. As neurosymbolic approaches often enhance interpretability, they could help to mitigate this uncertainty. In Section \ref{sub:nesyai}, the utilities of neurosymbolic AI for the DR challenge are demonstrated through some of the surveyed approaches, namely \textbf{Walking RDF and OWL} \cite{alshahrani2017neuro}, \textbf{PoLo} \cite{liu2021neural}, and \textbf{graph query embeddings} \cite{hamilton2018embedding}.

As neurosymbolic methods for reasoning on graphs are still growing in popularity, there are many outstanding challenges, common for reasoning on biomedical KGs \cite{li2022graph}, that are not addressed by the approaches discussed here. This includes the prediction of side effects \cite{deac2019drug, krix2022multigml}, polypharmacy side effects (involving interactions between two or more drugs) \cite{zitnik2018modeling, gema2023knowledge}, and protein function prediction \cite{xia2021geometric}. In addition to these, we suggest other prospective directions later in Section \ref{sec:prospective}.

\begin{figure*}[hbt!]
 \centering
 \small
 \begin{center}
  \input{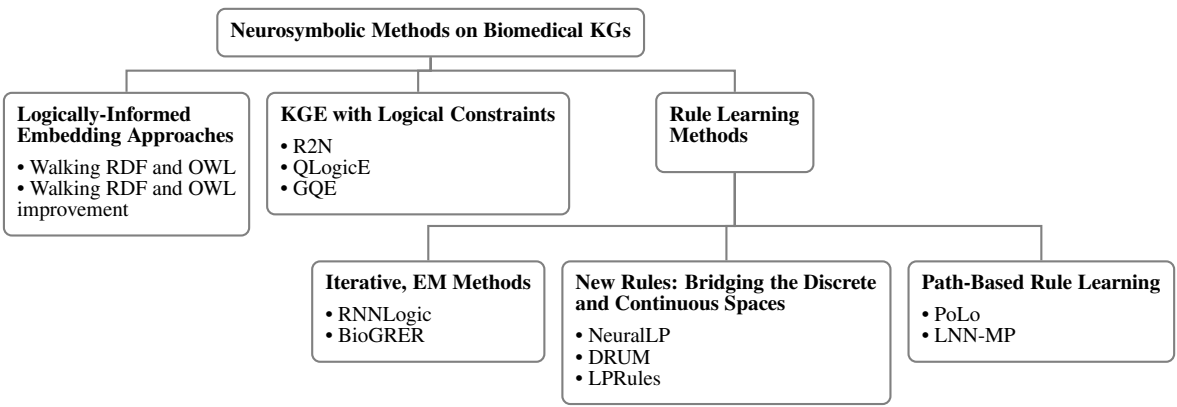}
  \begin{forest}
    mytree,
    [,phantom
        [\textbf{Neurosymbolic Methods on Biomedical KGs}
            [{\textbf{Logically-Informed}\\\textbf{Embedding Approaches}\\[0.5em]
                \textbullet \ Walking RDF and OWL\\  
                \textbullet \ Walking RDF and OWL\\improvement
            }]
            [{\textbf{KGE with Logical Constraints}\\[0.5em]
                \textbullet \ R2N\\
                \textbullet \ QLogicE\\
                \textbullet \ GQE 
            }]
            [{\textbf{Rule Learning}\\\textbf{Methods}\\[0.5em]},l sep+=20pt,
                [{\textbf{Iterative, EM Methods}\\[0.5em]
                    \textbullet \ RNNLogic\\
                    \textbullet \ BioGRER
                },
                edge path={
                  \noexpand\path [draw, \forestoption{edge}] (!u.parent anchor) -- +(0,-20pt) -| (.child anchor)\forestoption{edge label};
                }],
                [{\textbf{New Rules: Bridging the Discrete}\\\textbf{and Continuous Spaces}\\[0.5em]
                    \textbullet \ NeuralLP\\
                    \textbullet \ DRUM\\ 
                    \textbullet \ LPRules
                },
                edge path={
                  \noexpand\path [draw, \forestoption{edge}] (!u.parent anchor) -- +(0,-20pt) -| (.child anchor)\forestoption{edge label};
                }],
                [{\textbf{Path-Based Rule Learning}\\[0.5em]
                    \textbullet \ PoLo\\
                    \textbullet \ LNN-MP
                },
                edge path={
                  \noexpand\path [draw, \forestoption{edge}] (!u.parent anchor) -- +(0,-20pt) -| (.child anchor)\forestoption{edge label};
                }]
            ]
        ]
    ]
\end{forest}
 \end{center}
 \caption{Taxonomy of neurosymbolic approaches for reasoning on biomedical KGs \cite{delong2023neurosymbolic}.}
 \label{fig:taxonomy}
\end{figure*}

\section{Neurosymbolic Approaches on Biomedical KGs}
\label{sub:nesyai}

As established, we observe a dichotomy between methods for reasoning on graph structures using symbolic, rule-based methods (Section \ref{subsub:symbolic}) versus state-of-the-art KGE approaches (Section \ref{subsub:kge}). Recent studies in \emph{neurosymbolic AI} often combine aspects from deep learning and symbolic reasoning to mitigate their respective weaknesses and bridge such chasms \cite{tsamoura2020neural, garcez2020neurosymbolic, susskind2021neuro}. Throughout this survey, we discuss three major characteristics of these neurosymbolic approaches which benefit biomedical research:

\begin{enumerate}
    \item \textbf{Interpretability}: As mentioned, we often see a tradeoff between interpretability, which symbolic AI naturally possesses, and performance, in which black-box, deep learning methods seem to dominate. In particular, enforcing interpretability often comes with a drop in predictive performance \cite{dziugaite2020enforcing, molnar2022} and might lead to infeasible computational complexity \cite{bertsimas2019price, dziugaite2020enforcing}. Some of the surveyed neurosymbolic approaches, therefore, foster interpretability without sacrificing performance. As mentioned in the previous section, interpretability is valuable for biomedical applications: it could grant both healthcare providers and patients clarity as to how a decision was reached, as well as potentially expose reasoning processes derived from biases in the training data, such as imbalanced race or gender distributions \cite{olejarczyk2019patient, tat2020addressing, norori2021addressing}.
    \item \textbf{Integration of Domain Knowledge}: Biology is rich in patterned domain knowledge which is typically only implicitly represented in data-driven approaches. As mentioned in Section \ref{subsub:symbolic}, many biomedical ontologies exist to represent patterns of relationships and hierarchies. These include the Gene Ontology (GO) \cite{ashburner2000gene}, the Disease Ontology (DO) \cite{schriml2022human}, \& the Human Phenotype Ontology (HPO) \cite{kohler2021human}. There is also a widely accepted but manually designed classification system for macromolecules such as proteins \cite{lo2000scop, pearl2003cath} and their activities \cite{tipton2018brief}, as well as a developing system for classifying functional groups, or parts, on drugs \cite{guvench2016computational, peng2020motif}. The ability of some neurosymbolic approaches to integrate this knowledge into an otherwise data-driven approach is powerful. First, it bypasses the need for the model to learn patterns which are already known. Furthermore, it guides the learning toward more biologically accurate and relevant patterns.
    \item \textbf{Sparsity Handling}: As stated in Section \ref{sub:dichotomy}, knowledge within a biomedical KG is likely only a fraction of the what is discoverable. Furthermore, many clinical datasets have smaller proportions of certain genders and races \cite{tat2020addressing, puyol2021fairness} and infrequently contain instances of rare conditions \cite{zhao2018framework}. Data sparsity and imbalance are, therefore, major challenges within biomedical data. Through KG augmentation, some neurosymbolic approaches pose novel ways to address this.
\end{enumerate}

Within the following subsections, we refer back to these three characteristics and discuss the varied strategies by which different approaches cultivate them. Notably, to facilitate easy comparison, we group the approaches according to the taxonomy in Fig. \ref{fig:taxonomy}, based on categories from our previous survey \cite{delong2023neurosymbolic}. Finally, we highlight unique features of each approach that are particularly beneficial for biomedical prediction tasks. To facilitate accessibility, we curated a repository of available code for each work on GitHub\footnote{\url{https://github.com/NeSymGraphs}}.

\subsection{Logically-Informed Embedding Approaches}
\label{sub:logically_informed}

\begin{figure}[b!]
    \centering
    \input{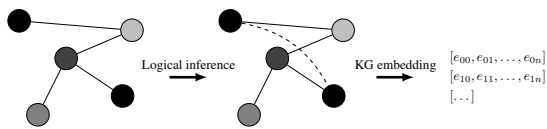}
    \scalebox{0.5}{\begin{tikzpicture}
    \tikzset{node_style/.style={shape=circle, draw=black, minimum size=6mm}}
    
    \node[node_style, fill=black] (A1) at (0, 0) {};
    \node[node_style, fill=lightgray] (B1) at (3, -0.25) {};
    \node[node_style, fill=darkgray] (C1) at (1.25, -1) {};
    \node[node_style, fill=gray] (D1) at (0.5, -2.5) {};
    \node[node_style, fill=black] (E1) at (2.75, -2) {};

    \path[-] (A1) edge node[] {} (B1);
    \path[-] (B1) edge node[] {} (C1);
    \path[-] (C1) edge node[] {} (D1);
    \path[-] (C1) edge node[] {} (E1);

    \node[node_style, fill=black, right = 5cm of A1] (A2) {};
    \node[node_style, fill=lightgray, right = 5cm of B1] (B2) {};
    \node[node_style, fill=darkgray, right = 5cm of C1] (C2) {};
    \node[node_style, fill=gray, right = 5cm of D1] (D2) {};
    \node[node_style, fill=black, minimum size=6mm, right = 5cm of E1] (E2) {};

    \path[-] (A2) edge node[] {} (B2);
    \path[-] (B2) edge node[] {} (C2);
    \path[-] (C2) edge node[] {} (D2);
    \path[-] (C2) edge node[] {} (E2);
    \path[-] (A2) edge[dashed, bend left=25] node[] {} (E2);

    \node[below right = 1.25cm and 5.5cm of A2, text centered, text height=0cm] (O) {
        $\begin{aligned}
             &[e_{00}, e_{01}, \dots, e_{0n}] \\
             &[e_{10}, e_{11}, \dots, e_{1n}] \\
             &[\dots ]
        \end{aligned}$
    };

    \path[-latex, line width=2pt] (4, -1.5) edge node[above] {Logical inference} (5, -1.5);
    \path[-latex, line width=2pt] (9.5, -1.5) edge node[above] {KG embedding} (10.5, -1.5);
\end{tikzpicture}}
    \caption{\textbf{Logically-Informed Embedding Approaches} \cite{delong2023neurosymbolic}.}
    \label{fig:modular_two_step}
\end{figure}

The most straightforward way to combine the benefits of symbolic and KGE methods is to use them in succession. Some neurosymbolic approaches on graphs modularize the two, then feed the results from the former into the latter. Because symbolic approaches are often based on expert-defined rules, they can be viewed as methods to extend the ground truth and are therefore used as a preliminary step (the logic module) which feeds into the KGE step (the neural module). One approach \cite{alshahrani2017neuro}, for example, employed an ontological reasoner to infer new edges in their KG, followed by a common KGE method called DeepWalk \cite{perozzi2014deepwalk}. Consequently, the generated embeddings contained richer information as they were based on an augmented KG. Their method, \textbf{Walking RDF and OWL}, is a specific example of the more general pattern illustrated in Fig. \ref{fig:modular_two_step}. Using these semantic-aware embeddings, they performed link prediction for a DR task. Not only did prediction performance improve with the augmented KG, but their model could also accurately predict other links, like side effects. Another group \cite{agibetov2018fast} improved upon this approach with an alternative, log-linear embedding method. 

These studies effectively demonstrate several ways in which biomedical domain knowledge can be integrated into a KGE approach. First, Alshahrani \textit{et al.} develop their own KG from various sources, including both evidence-based interactions from databases and higher-level relationship patterns from biomedical ontologies, including the GO, DO, and HPO. By mapping individual biomedical entities such as proteins and drugs to their respective classifications in these ontologies, they expand the dimensionality of the KG through those higher-level relationships. Furthermore, while \textbf{Walking RDF and OWL} uses an ontological reasoner nonspecific to biology, this general approach presents the ideal opportunity to use biomedical ontologies or rules in the KG augmentation step.

The KG augmentation step also helps with sparsity handling. As mentioned, some clinical datasets have fewer instances of rare diseases, so there will be less known indications, side effects, and processes associated with those diseases. On the molecular level, we also see disparities in the volume of data surrounding certain relations; particular BPs are especially difficult to study as the involved macromolecules are harder to isolate \textit{in vitro} or degrade easily \cite{lee2017simple}. It is possible that, while there is expert knowledge about the way these rare relations occur, a purely data-driven model might not discover these patterns due to imbalance in the training data. If the KG is initially augmented using expert-defined or domain-specific rules, this may increase the number of these rare relations on which the KGE model is trained. As their application is DR, this concept is particularly important for the discovery of novel treatments: the prediction of indications for uncommon diseases or drugs is likely more impactful than that for more common ones.

\subsection{KGE with Logical Constraints}
\label{sub:constraints}

In contrast to integrating biomedical knowledge through KG augmentation, other approaches use it to impose logical or rule-based constraints to the KGE pipeline. In this case, the main focus is typically on training the KGE method, but rules limit the scope of predictions possible. These approaches fundamentally differ based on where logical constraints are imposed in the learning pipeline.

For example, logical constraints can be applied directly to the latent space embeddings, as shown in Figure \ref{fig:constraints_embedding}. The \textbf{Relational Reasoning Network (R2N)} \cite{marra2021learning}, for instance, uses structural relational information as constraints on GNN training. This transforms the latent space embeddings in a way that is dependent upon each node’s neighbors in the original space so that learned KGEs also encode information about each node's surrounding graph topology. This is ideal for biomedical networks as many interactions are dependent upon some adjacent interaction. For example, a protein called a GTPase alters a molecule called Guanosine triphosphate (GTP), and this reaction is highly involved in the cell death BP \cite{he2021roles}. However, a GTPase is regulated by interaction with another protein, the GTPase-activating protein \cite{he2021roles}. Accounting for neighboring graph topology, therefore, helps capture such relationships. Consequently, the \textbf{R2N} encodes high-level relationships like the approaches in Section \ref{sub:logically_informed}, but in \textbf{R2N}, the corresponding truth values for these relationships are incrementally adjusted. Furthermore, while such relationships could be encoded into \textbf{R2N} explicitly like they were in \textbf{Walking RDF and OWL}, their particular use on the UMLS KG relies, instead, on implicit logic rules which are not made available to the user. Unfortunately, this is a missed opportunity regarding characteristics 1 and 2 described at the start of Section \ref{sub:nesyai}. We see the same traits in the \textbf{Quantum Logic Empowered Embedding (QLogicE)} \cite{chen2022qlogice} method which computes logic embeddings and KGE in parallel. The scoring function for each possible triple is a weighted sum of the scores for the two embedding methods, and the loss is determined as the weighted sum of the loss of the two embedding methods. Like \textbf{R2N}, high-level relationships between general concepts are determined in the \textbf{QLogicE} model through the logic embedding module, but they are not available for the user to explore. While \textbf{QLogicE} achieves competitive performance on the UMLS KG, the lack of interpretability limits the utility in biomedicine.

Also like \textbf{R2N}, another study \cite{hamilton2018embedding} provides a different way to incorporate logical constraints encoding dependencies between adjacent edges. Here, they motivate the need for \textit{conjunctive} queries on a biomedical KG through the DR task: they pose the idea that one might query for a drug that simultaneously treats two conditions via a commonly involved protein. This query involves the conjunction of three concepts: that a drug targets some protein, and that the protein is associated with both diseases. Here, we see again that the sought indications rely on the adjacent relationships between the protein and the diseases. Their method, \textbf{graph query embeddings (GQE)}, combines logic and node embedding spaces in a way that the logic embeddings represent a query. It does so by learning both node embeddings and trainable parameter matrices which represent logical operators. Together, they use these to compute \textit{query embeddings}. Thereafter, the likelihood that a set of nodes satisfy a query is calculated as the cosine similarity between the query embeddings and the respective node embeddings. Similarly to \textbf{R2N}, \textbf{GQE} imposes logical constraints onto the embedding space. However, \textbf{GQE} is arguably better at integrating domain knowledge because the structure of the conjunctive query is a more task-specific way to make predictions.

In general, while imposing logical constraints allows multiple ways to incorporate domain-specific information, it does not necessarily add any aspects of interpretability. To take full advantage of the neurosymbolic concept of hybridizing performance with interpretability, the logic and neural components of an architecture should ideally be combined in a way that user-level explanations are generated. In the next section, we review several approaches which do just that.

\begin{figure}[t!]
    \centering
    \input{figures/tikz_definitions.tex}
    \scalebox{0.5}
    {\begin{tikzpicture}
    \def\graphsep{5cm}
    \tikzset{node_style/.style={shape=circle, draw=black, minimum size=6mm}}
    
    \node[node_style, fill=black] (A1) at (0, 0) {};
    \node[node_style, fill=lightgray] (B1) at (3, -0.25) {};
    \node[node_style, fill=darkgray] (C1) at (1.25, -1) {};
    \node[node_style, fill=gray] (D1) at (0.5, -2.5) {};
    \node[node_style, fill=black] (E1) at (2.75, -2) {};

    \path[-] (A1) edge node[] {} (B1);
    \path[-] (B1) edge node[] {} (C1);
    \path[-] (C1) edge node[] {} (D1);
    \path[-] (C1) edge node[] {} (E1);

    \node[below right = 1.25cm and 6cm of A1, text centered, text height=0cm] (O) {
        $\begin{aligned}
             &[e_{00}, e_{01}, \dots, e_{0n}] \\
             &[e_{10}, e_{11}, \dots, e_{1n}] \\
             &[\dots ]
        \end{aligned}$
    };

    \path[-latex, line width=2pt] (4.25, -1.5) edge node[above] (LI) {KG embedding} (5.25, -1.5);

    \node[above = 1cm of LI] {Constraints};
    \path[-latex, line width=2pt] (4.75, 0) edge node[] {} (4.75, -0.75);
\end{tikzpicture}}
    \caption{\textbf{KGE with Logical Constraints} \cite{delong2023neurosymbolic}.}
    \label{fig:constraints_embedding}
\end{figure}

\subsection{Rule Learning Methods}
\label{sub:rule-learning}

In the previous sections, all of the described methodologies use predefined rules. However, many other approaches attempt to \textit{learn} rules and rule confidences for KGC.

\subsubsection{Iterative, EM Methods}
\label{subsub:iterative}

Many rule-learning methods claim to take an Expectation-Maximization algorithm (EM) based approach, alternating between two modules whose outcomes inform each other. The EM algorithm is used to estimate the maximum likelihood of model parameters for situations in which there is incomplete data \cite{gupta2011theory, ng2012algorithm}. KGC, then, is an ideal and obvious setting for such methods. The iterative nature of the EM algorithm between the prediction of missing components and the optimization of model parameters to account for such predictions is the core idea here. Similar to approaches described in Section \ref{sub:logically_informed}, these methods use two complementary modules to inform one another, but the logic module is now dynamic, adjusting the rule base at each iteration. This bidirectional iteration is shown in Fig. \ref{fig:em_methods}. These methods vary, however, in their interpretations of how an EM-based algorithm is executed. 

\begin{figure}[t!]
    \centering
    \input{figures/tikz_definitions.tex}
    \scalebox{0.5}{\begin{tikzpicture}
    \def\graphsepright{5cm}
    \def\graphsepright2{10cm}
    \def\graphsepabove{2cm}
    \tikzset{node_style/.style={shape=circle, draw=black, minimum size=6mm}}
    
    \node[node_style, fill=black] (A1) at (0, 0) {};
    \node[node_style, fill=lightgray] (B1) at (3, -0.25) {};
    \node[node_style, fill=darkgray] (C1) at (1.25, -1) {};
    \node[node_style, fill=gray] (D1) at (0.5, -2.5) {};
    \node[node_style, fill=black] (E1) at (2.75, -2) {};

    \path[-] (A1) edge node[] {} (B1);
    \path[-] (B1) edge node[] {} (C1);
    \path[-] (C1) edge node[] {} (D1);
    \path[-] (C1) edge node[] {} (E1);

    \node[below right = 2.5cm and 4.75cm of A1, text centered, text height=0cm] (O) {
        $\begin{aligned}
             &[e_{00}, e_{01}, \dots, e_{0n}] \\
             &[e_{10}, e_{11}, \dots, e_{1n}] \\
             &[\dots ]
        \end{aligned}$
    };

    \node[above = 2cm of O] (Rule) {
        $\begin{aligned}
             &(p_1) && \text{Rule 1}\\
             &(p_2) &&\text{Rule 2}\\
             & && \cdots \\ 
             & && \text{new rules?}
        \end{aligned}$
    };

    \node[node_style, fill=black, right = 9.5cm of A1] (A3) {};
    \node[node_style, fill=lightgray, right = 9.5cm of B1] (B3) {};
    \node[node_style, fill=darkgray, right = 9.5cm of C1] (C3) {};
    \node[node_style, fill=gray, right = 9.5cm of D1] (D3) {};
    \node[node_style, fill=black, minimum size=6mm, right = 9.5cm of E1] (E3) {};

    \path[-] (A3) edge node[] {} (B3);
    \path[-] (B3) edge node[] {} (C3);
    \path[-] (C3) edge node[] {} (D3);
    \path[-] (C3) edge node[] {} (E3);
    \path[-] (A3) edge[dashed, bend left=25] node[] {} (E3);
    \path[-] (B3) edge[dashed, bend left=15] node[] {} (D3);

    \path[-latex, line width=2pt] (3.75, -0.5) edge node[sloped, anchor=center, above=1mm] {} (4.75, 0.5);
    \path[-latex, line width=2pt] (3.75, -2) edge node[sloped, anchor=center, below=1mm] {} (4.75, -3);
    \path[-latex, line width=2pt] (8, 0.5) edge node[sloped, anchor=center, above=1mm] {Logical inference} (9, -0.5);
    \path[-latex, line width=2pt] (8, -3) edge node[sloped, anchor=center, below=1mm] {Decoding} (9, -2);

    \path[-latex, line width=2pt] (11, -3) edge[bend left] node[sloped, anchor=center, below=1mm] {E-step} (7, -3.5);
    \path[-latex, line width=2pt] (11, 0.5) edge[bend right] node[sloped, anchor=center, above=1mm] {M-step} (7, 1.5);
\end{tikzpicture}}
    \caption{\textbf{Iterative, EM Algorithm Methods} \cite{delong2023neurosymbolic}.}
    \label{fig:em_methods}
\end{figure}

\textbf{RNNLogic} \cite{qu2020rnnlogic}, unlike other KGC methods, employs an EM-based algorithm without any sort of KGE. Instead, \textbf{RNNLogic} iterates between training a rule generator (the ``M-step'') and an inference step (the ``E-step''). The two inform one another to reduce the search space of possible logical rules. In contrast, Zhao \textit{et al.}'s method, \textbf{Biomedical KG refinement with Embedding and Rules (BioGRER)} \cite{zhao2020biomedical} incorporates a KGE method. During the ``E-step'', logical rules are used to infer novel triples, and those inferred triples are used to train the KGE method. Thereafter, the predicted triples from the KGE method are used to inform and update the corresponding rule weights in the logic module as the ``M-step''.

Both methods foster interpretability in a way that the previous approaches do not: they make both the rules and the corresponding weights available to the end-user. This means that the user can see which reasoning processes contributed to the prediction process and which were most influential. Zhao \textit{et al.}, who used kg-covid-19 (see Table \ref{tab:kgs}), give specific examples of learned rules, such as, 
\(
\textit{interacts\_with}(x, y) \Leftrightarrow \textit{interacts\_with}(y, x). 
\)
Because \textbf{BioGRER} uses generic \textit{rule models} as templates for learned rules, a user can encode domain-specific knowledge into these rule models. For example, the transitive rule written as 
\(
\textit{r$_i$}(x, y) \land \textit{r$_i$}(y, z) \rightarrow \textit{r$_i$}(x, z) 
\)
may not be as useful in the biomedical context. Consider, for instance, the \textit{binds} relation between two proteins, indicating that they physically and chemically connect. Two \textit{binds} relations may constitute adjacent edges but not indicate transitivity in that sense. Transitivity, however, can be extended to two different but adjacent relations which result in a third, overarching relation. Zhao \textit{et al.} give the clinical example:
\(
\textit{tributary\_of}(x, y) \land \textit{drains}(y, z) \rightarrow \textit{part\_of}(x, z) 
\).
\textbf{BioGRER}, therefore, uses an extended version of the transitive rule as it suits the application domain better:
\(
\textit{r$_i$}(x, y) \land \textit{r$_j$}(y, z) \rightarrow \textit{r$_k$}(x, z) 
\).
Thus, \textbf{BioGRER} satisfies characteristics 1 and 2 at the start of Section \ref{sub:nesyai}. In fact, both \textbf{RNNLogic} and \textbf{BioGRER} satisfy the third characteristic of sparsity handling in the same way that the approaches from Section \ref{sub:logically_informed} do. Because they use rule-based inference in their logic modules, they can increase the instances of rare relations in the data on which the neural module is subsequently trained. Within the next section, we survey approaches which take the rule-learning strategy a step further, generating new rules for the logic module.

\subsubsection{New Rules: Bridging the Discrete and Continuous Spaces}

The previous methods assigned weights to rules which were mined or constructed according to a template, but other approaches use the learning process to drive the construction of \textit{new} rules. To learn and generate novel rules, an algorithm must learn both structural information in a discrete space and respective weight parameters in a continuous space. The creation of a method which can do both and train in an end-to-end fashion is inherently difficult \cite{yang2017differentiable, sadeghian2019drum}. The benefit, though, is that the resultant rules are more descriptive of the reasoning process, thereby enhancing interpretability and improving upon point 1 from the start of Section \ref{sub:nesyai}.

\textbf{LPRules} \cite{dash2021lprules}, for example, repeatedly augments a small starter pool of candidate rules, optimizing predictive capability over iterations. By doing so, they minimize the search space in comparison to enumerating and testing all possible rules. Alternatively, \textbf{Neural LP} \cite{yang2017differentiable} avoids rule enumeration by learning confidence values for every relation involved in every rule body, even accounting for differences in the varying lengths. This makes the rule bodies dynamic with the learning process. \textbf{DRUM} \cite{sadeghian2019drum}, is a refined version of \textbf{Neural LP} in which the authors point out that \textbf{Neural LP} learns high confidences for incorrect rules. \textbf{DRUM} additionally encodes information about how feasibly two relations can co-exist within a rule body. This is particularly interesting for capturing domain-specific knowledge. For example, on the molecular scale, certain functions can be highly specialized to a subcellular location like the nucleus or mitochondria \cite{lu2005go}; these functions are unlikely to co-occur in a chained pattern. However, while learned rules are, in theory, more expressive, there is a major caveat to these methods. \textbf{Neural LP} and \textbf{DRUM} only train on positive examples and have yet to be tried on a dataset with negative ones. A lack of negative examples in the training data increases the risk of false positive predictions, which, in contexts such as such as DR, could lead to a waste of time, resources, or potentially even health hazards. 

\subsubsection{Path-based Rule Learning}
\label{subsub:pathbased}

Most of the previous approaches treat rules as statements involving individual relation types. Path-based methods, however, learn and make inferences from \textit{chains} of edges or relations. Path-based rules are not only interpretable but often more expressive, as they facilitate understanding of relationships between nodes that are several hops away from each other, a trait which we call \textit{long-range dependency} \cite{liu2021neural}. 

Long-range dependency is particularly important in biology in which we frequently see chains of interactions between macromolecules. For instance, molecular signaling cascades describe a chain in which the interaction between two proteins is dependent upon one's contact with another molecule upstream \cite{chang2001mammalian, crino2016mtor}. In fact, there are several databases storing such pathway information \cite{d2013pathway}. This is why one study, called \textbf{Policy-guided walks with Logical Rules (PoLo)} \cite{liu2021neural}, uses a path-based neurosymbolic method for DR. By exploring indirect drug indications that might operate through other entities, such as genes, they expand the set of novel discoveries possible. \textbf{PoLo} encodes general path patterns, or \textit{metapaths}, as logical rules with pre-computed confidence scores. While metapaths can be manually selected from expert knowledge, they can also be mined, such as in a follow-up study \cite{drance2021neuro}. Therefore, these path-based methods are not only interpretable, but they also hold potential to include domain knowledge.

Sometimes, less frequent path sequences might be less relevant to the end-goal. Although molecular components interact in complex, multi-relational webs, we still see both physical \cite{lu2005go} and functional compartmentalization \cite{ashburner2000gene}. For example, if a study is searching for a drug which aids in the cellular detoxification process, the paths involved in cell replication might be less relevant. Another group \cite{sen2021combining} addresses such a situation with their \textbf{LNN-MP} method by generating all possible length-$k$ path-based rules, then using them on KGEs pre-trained with bias toward more prevalent relation sequences. By doing so, \textbf{LNN-MP} learns to generate rules most relevant to the paths present in the KG.

Of the three major categories, the rule-learning methods most often cover the three characteristics mentioned at the start of Section \ref{sub:nesyai}: their rule production adds interpretability to otherwise black-box models, some admit the use of domain-specific rules, and their use of logical inference can augment the instances of rare relations. Next, we explore the prospective opportunities  to exploit these characteristics.

\section{Prospective Directions}
\label{sec:prospective}

As mentioned, there are plenty of technical and practical areas yet to be fully explored, especially in biomedicine. Here, we suggest a number of prospective directions, which we hope will cultivate a greater interest in this domain.

\subsection{Mechanism of Action Discovery}

As mentioned in Section \ref{subsub:pathbased}, many molecular interactions rely on long-range dependencies. In particular, some drug indications rely on a chain of events \cite{schenone2013target}. Unfortunately, the path by which a drug compound induces a therapeutic effect, which we call the mechanism of action (MOA) \cite{schenone2013target}, is not always straightforward or understood. Furthermore, cellular interactions can be nonspecific, so side effects may occur as a result \cite{schenone2013target}. By using the rules learned by one of the path-based methods from Section \ref{subsub:pathbased}, one could potentially uncover unknown MOAs of drugs. Revealing the MOA would facilitate a deeper understanding of how directly or indirectly a given drug achieves some therapeutic effect,  what additional therapeutic or adverse effects can be expected from usage of said compound, and how such effects may vary across cell and tissue types. 

\subsection{Spatiotemporal Reasoning}

With respect to understanding differences across cell and tissue types, biomedical networks also frequently have spatiotemporal dependencies, as mentioned in Section \ref{sub:biology}. Previous approaches for spatiotemporal-conscious reasoning on graph structures were based on deep learning architectures \cite{yu2017spatio, chen2018gc, li2019hybrid}. However, we have yet to observe many neurosymbolic approaches. For example, methods from Section \ref{sub:rule-learning} could learn rules which describe the relationships between the various edge types and time. Alternatively, rules could be learned for each specific time range or spatial location. As mentioned, the types and abundance of certain proteins vary across tissue types, so the levels to which proteins affect one another are spatially dependent \cite{buccitelli2020mrnas}. Learning rules specific to tissue types could better inform clinical researchers as to how processes differ between tissues.

Alternatively, instead of learning rules, methods like those in Section \ref{sub:constraints} could encode time- or space-dependent rules as constraints. For example, in a biological network, the half-lives of molecules such as messenger ribonucleic acid not only act as a time limit on potential interactions but also determine the levels to which various proteins are expressed, and therefore the extent to which those proteins can have effects on other players in the network \cite{mauger2019mrna, buccitelli2020mrnas}. Such constraints can more realistically model spatiotemporal dependencies by adding logic or domain-specific rules.

\subsection{Conditional or Interdependent Edge Types}

In some cases, certain relations or edge types might be dependent upon one another in a way that one edge does not occur unless other edges exist. An example of this can be found in the case of molecular signaling cascades in which one protein will not interact with another unless contact is made with another protein or molecule upstream \cite{chang2001mammalian, crino2016mtor}. Alternatively, such dependence might also exist in a way that more frequent occurrences of a given relation have some effect on another relation, such as increasing or decreasing the frequency. Taking again the signaling cascade example, the more an inhibitor interacts with some protein, the less that protein might be able to take part in other interactions \cite{gubb2010protease, pleschka2001influenza}. While such dependencies and conditional scenarios are not unique to neurosymbolic reasoning, it might be ideal for addressing them. With the incorporation of rules, neurosymbolic methods could provide novel ways to encode dependency information between relation types or individual relations. Potentially, methods from Section \ref{sub:constraints}, which use logic to assess the probability that an edge is implied from its adjacent edges, could be ideal for interdependent edge types. Furthermore, since methods from Section \ref{sub:rule-learning} learn confidences for rules, there is also potential to learn weights between rules that coexist or influence one another. Neurosymbolic methods for graphs could, therefore, provide unique ways to interpret such dynamic biomedical KGs and the relationships between interdependent molecular interactions.

\subsection{Few Shot Learning}

As already mentioned, methods such as the approaches in Sections \ref{sub:logically_informed} and \ref{sub:rule-learning} increase the number of rare relation types through KG augmentation. Such approaches could, therefore, be used as a solution for few shot learning problems. In few shot learning, there exist a small number of instances of some class within the training data \cite{yue2020interventional}. This could promote research which is more inclusive of rare diseases or BPs.

\section{Conclusion}

In this survey, we focus on how the unique traits of neurosymbolic methods for graphs are fitting for a multitude of challenges in biomedicine, condensing some of the concepts covered by our previous survey \cite{delong2023neurosymbolic}. We provide a methodological taxonomy by which to classify these approaches, and we discuss the ways in which they facilitate interpretability, knowledge-integration, and sparsity handling within the context of biomedical KGC. Additionally, we curated a repository of available code for each work on GitHub\footnote{\url{https://github.com/NeSymGraphs}}. Finally, with hopes to guide future research, we propose prospective biomedical directions toward which neurosymbolic methods for reasoning on graphs might steer.

\section*{Acknowledgements}

LND is funded by the University of Edinburgh (UoE) Informatics Graduate School through the Global Informatics Scholarship. RFM is funded by the UoE Institute for Academic Development through the Principal's Career Development PhD Scholarship. FNCS is supported by the United Kingdom Research and Innovation (grant EP/S02431X/1), UKRI Centre for Doctoral Training in Biomedical AI at the UoE, School of Informatics. For the purpose of open access, the author has applied a creative commons attribution (CC BY) licence to any author accepted manuscript version arising. We thank these institutions for their support.


\bibliography{references}
\bibliographystyle{icml2023}

\newpage
\appendix
\onecolumn


\end{document}